\RequirePackage{scrlfile}
\PreventPackageFromLoading{subfig}
\documentclass[letterpaper, 10 pt, conference]{ieeeconf}  

\IEEEoverridecommandlockouts                   
\usepackage[table,xcdraw]{xcolor}
\usepackage{siunitx}
\usepackage{graphics} 
\usepackage{epsfig} 
\usepackage{float}
\usepackage{mathptmx} 
\usepackage{times} 
\usepackage{amsmath} 
\usepackage{amssymb}  
\usepackage{subcaption}
\usepackage{svg}
\usepackage{hyperref}
\usepackage{pgf,tikz}
\usepackage{mathrsfs}
\usetikzlibrary{arrows}
\usepackage{xcolor}
\usepackage{stackengine,scalerel}
\usepackage[pdf]{pstricks}
\usepackage[table,xcdraw]{xcolor}
\usepackage{commath}
\usepackage{subfig}

\usepackage[T1]{fontenc}
\usepackage[utf8]{inputenc}
\usepackage[none]{hyphenat}

\stackMath

\title{\LARGE \bf An LSTM Network for Real-Time Odometry Estimation 
}
 
\author{Michelle Valente$^{1}$, Cyril Joly$^{1}$ and Arnaud de La Fortelle$^{1}$
\thanks{*This work was supported by the International \emph{Chair Drive for All} with sponsors PSA Peugeot Citroën - Safran - Valeo}
\thanks{$^{1}$Michelle Valente, Cyril Joly and Arnaud de la Fortelle are with Center for Robotics, MINES ParisTech, PSL Research University, 60 boulevard Saint Michel, 75006 Paris, France. {\tt\small \{michelle.valente, cyril.joly, arnaud.de\_la\_fortelle\}@mines-paristech.fr}}%
}%

\begin{document}
\maketitle

\begin{abstract}

The use of 2D laser scanners is attractive for the autonomous driving industry because of its accuracy, light-weight and low-cost. However, since only a 2D slice of the surrounding environment is detected at each scan, it is a challenge to execute important tasks such as the localization of the vehicle. In this paper we present a novel framework that explores the use of deep Recurrent Convolutional Neural Networks (RCNN) for odometry estimation using only 2D laser scanners. The application of RCNNs provides the tools to not only extract the features of the laser scanner data using Convolutional Neural Networks (CNNs), but in addition it models the possible connections among consecutive scans using the Long Short-Term Memory (LSTM) Recurrent Neural Network. Results on a real road dataset show that the method can run in real-time without using GPU acceleration and have competitive performance compared to other methods, being an interesting approach that could complement traditional localization systems. 

\end{abstract}

\section{Introduction}

In the last years, Deep Learning methods have been receiving attention in the field of autonomous driving. The most common applications use cameras or laser scanners for tasks such as obstacle detection \cite{detection_camera}\cite{detection_lidar} and classification \cite{classification}. These tasks are mainly in the field of environment understanding and mapping. However, besides the need to understand the environment, the localization of the vehicle is still a challenging process and it has not yet been extensively explored by machine learning techniques. 

To navigate unknown environments, methods known as Simultaneous Localization and Mapping (SLAM) allow to localize the vehicle while concurrently mapping the environment. Different types of sensors can be used for the mapping, such as cameras, laser scanners and radars. The use of 3D laser scanners has become very popular for autonomous driving, however its cost is still a major drawback for automobile manufacturers to maintain a reasonable price for the vehicles. Moreover, the amount of data provided by 3D laser scanners requires large computational resources. Considering this, we chose to focus on a solution for vehicles equipped with low cost 2D laser scanners. We propose a complete neural network method to localize a vehicle using only as input sequences of data acquired from 2D laser scanners. 

\begin{figure}[H]
	\centering
	\includegraphics[width=1.0\linewidth]{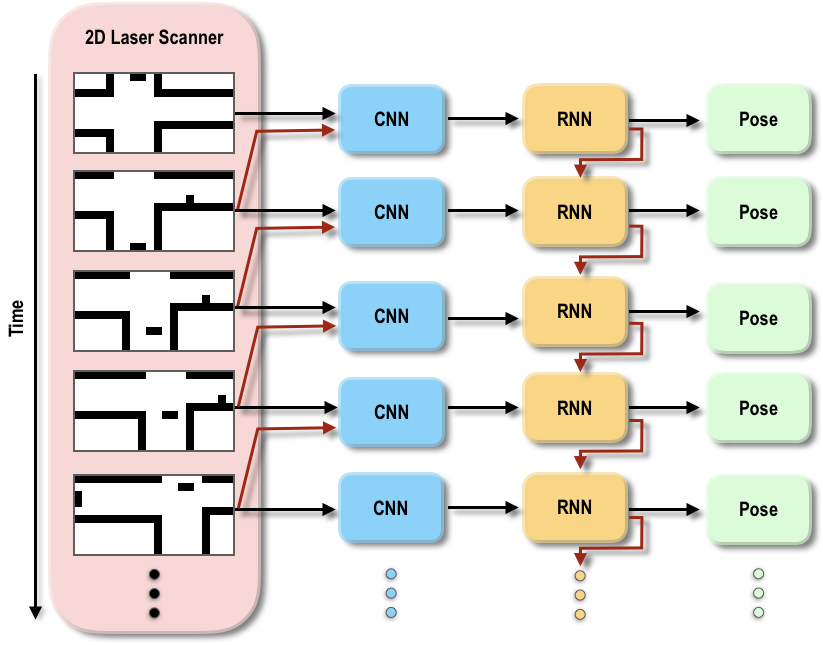}
	\caption{Overview of the proposed system. The complete Recurrent Convolutional Neural Network takes a sequence of 2D laser scanner measurements as input, learns its features by a sequence of CNNs, which are used by the RNN to estimate the poses of vehicle. The output is a 2D pose of the vehicle composed by two values, one for translation and another one for rotation.}
	\label{fig:System}
\end{figure}

Acquisition of a large amount of data is still one of the main challenges for the application of Deep Learning methods. Most of the datasets for autonomous driving with a significant number of sequences and good ground truth for localization are equipped with 3D laser scanners. For this reason, we use the KITTI \cite{kitti} odometry dataset and we simulate a 2D laser scanner by extracting one of the Velodyne layers. Although it is not a realistic position for the sensor, we can still use it as a proof of concept for estimating localization using this type of method. 

The proposed network is in the format of a Recurrent Convolutional Neural Network, as shown in Figure \ref{fig:System}. The main idea is to use a sequence of CNNs in order to extract the features between two sequential laser scanners. Sequentially, these features are the input of a Recurrent Neural Network (RNN), more specifically a Long Short-Term Memory (LSTM) RNN, which learns how to estimate the pose of the vehicle. 

The remainder of the paper is organized as follows. First, we present the related work in Section~\ref{relatedwork.sec}; the proposed method and the design of the network is presented in Section~\ref{method.sec}; experimental results are presented in Section~\ref{results.sec}; finally conclusion and perspectives are given in Section~\ref{conclusion.sec}.

\section{Related Work}\label{relatedwork.sec}

Deep learning research has presented notable results in different type of applications, among the most commons are object detection and classification using images. The success of these applications are mainly due to the fact that a large amount of datasets has become available in the last years. In the context of intelligent vehicles, interesting work has been developed in several different fields: trajectory prediction \cite{trajectory}\cite{trajectory2}, mapping \cite{mapping}, control \cite{control} and even end-to-end approaches \cite{endtoend}\cite{endtoend2}, where the car is controlled completely by a Deep Learning module.

At the same time, localization, which is still a very challenging problem for robotic systems, is not yet well explored by deep learning methods. Most of the proposed approaches are based on Visual Odometry (VO) \cite{VO}, which consists in using camera images for odometry estimation. The classic methods \cite{orb}\cite{svo} rely on finding geometry constraints from the images to estimate the camera's motion; while recent work tries to re-formulate this classic approach as a deep learning based VO, that potentially could deal with challenging environments and camera parameters difficulties. First, PoseNet \cite{posenet} proposed the use of CNNs to deal with the SLAM problem as a pose regression task by estimating the 6-DoF pose using only RGB images. Wang et al. \cite{deepvo} then introduced the DeepVO method, which uses RCNNs with the same goal. The same authors also presented the method UndeepVO \cite{undeepvo}, which proposes an unsupervised deep learning method to estimate the pose of a monocular camera. However, the classic VO methods still outperform deep learning based methods published to this date, considering the accuracy in the pose estimation.

The use of laser scanners is also popular for pose estimation in robotic systems. The most classic method consists in matching two point clouds to estimate the transformation between them by minimizing the matching error, this solution is known as Iterative Closest Point (ICP) \cite{icp}. More complex and robust methods to achieve the same goal have been proposed since then, like LOAM \cite{loam}, which runs two different algorithms in parallel to achieve real time processing. One algorithm can run faster to obtain a low fidelity odometry, and another one slower for a more precise matching. One of the main challenges for this type of localization methods is the error caused by moving obstacles or lack of information because of occlusions. For this reason, some localization solutions \cite{gridmatching}\cite{imls} are based on detecting what is static in the environment to only use this information for the scan matching process. 

The application of deep learning techniques for this purpose using laser scanners is still considered as a new challenge and only few papers have addressed it. The use of this type of technique can eliminate the computational time problem found in most of the localization methods, and even be able to deal with the moving obstacles difficulties. A network that is properly trained could be able to distinguish what are the best matching features (static obstacles) from the moving obstacles that can cause pose estimation errors.  Nicolai et al. \cite{laser1} were the first to propose to apply 3D laser scanner data in CNNs to estimate odometry. They presented an interesting approach that provided a reasonable estimation of odometry, however still not competitive with the efficiency of state-of-the-art scan matching methods. Later, Velas et al. \cite{laser3} presented another approach for using CNNs with 3D laser scanners for IMU assisted odometry. Their results were able to get high precision and close results compared to state-of-the-art methods, such as LOAM \cite{loam}, for translation, however the method is not able to estimate rotation with sufficient precision. Considering their results, the authors propose that their method could be used as a translation estimator and use together an Inertial Measurement Unit (IMU) to obtain the rotation. Another drawback is that, according to the KITTI benchmark, even using CNNs the method is slower than LOAM. 

In our work we chose to use as the sensor only a 2D laser scanner, which could reduce considerably the price of future intelligent vehicles. At this moment, Li et al. \cite{laser2} were the only authors to use this kind of sensor to estimate odometry by the use of a deep learning approach. They propose a network to perform scan matching and loop closure using CNNs. Although the loop closure networks shows good accuracy, the scan matching results are still very inaccurate compared to classic methods.  

Based on the idea presented in \cite{laser2}, we propose a new solution for 2D laser-based odometry estimation using deep learning networks. We explore the use of a RNN along with CNNs to learn temporal features in order to improve the odometry results. We also propose a new configuration for the CNNs where we were able to achieve better results. Finally, we explore this solution in outdoor environments, training and testing it with the KITTI \cite{kitti} dataset, which contains sequences different of type of scenarios.

\section{Method}\label{method.sec}

The proposed approach consists in finding the vehicle displacement by estimating the transformation between a sequence of 2D laser scanner acquisitions. From two consecutive observations, where each observation is a $\ang{360}$ set of points measured during one laser rotation, the network predicts the transformation $T = [\Delta{d}, \Delta{\theta}]$, which represents the travelling distance $\Delta{d}$ and the orientation $\Delta{\theta}$ between two consecutive laser scans $(s_{t-1}, s_{t})$ . We only consider the 2D displacement of the vehicle, since we are relying only on a 2D sensor. Therefore, the goal is to learn the optimal function $g(.)$, which maps $(s_{t-1}, s_{t})$ to $T$ at time $t$:
\begin{equation}
    T_t = g(s_{t-1}, s_{t})
\label{learning_function}
\end{equation}

Once we learn these parameters, we can obtain the 2D pose $(x_t,y_t,\theta_t)$ of the vehicle in time $t$  as follow:

\begin{equation}
\begin{split}
    x_t& = x_{t-1} + \Delta{d} \sin{(\Delta{\theta})} \\
    y_t& = y_{t-1} + \Delta{d} \cos{(\Delta{\theta})} \\
    \theta_t& = \theta_{t-1} + \Delta{\theta}
\end{split}
\end{equation}

In this way, we can accumulate the local poses of the vehicle and estimate the global position of the vehicle at any time $t$. Since the algorithm does not perform any sort of loop closure, drift can be also accumulated, thus reducing the accuracy of the vehicle's localization. 

The following subsections will present in details the proposed method. First, we show the raw data of the laser scanner is encoded. Sequentially, we present the configuration of the network and the specifics of the training process. 

\subsection{Data encoding}\label{sub.encoding}

\begin{figure}
	\centering
	\includegraphics[width=1.0\linewidth]{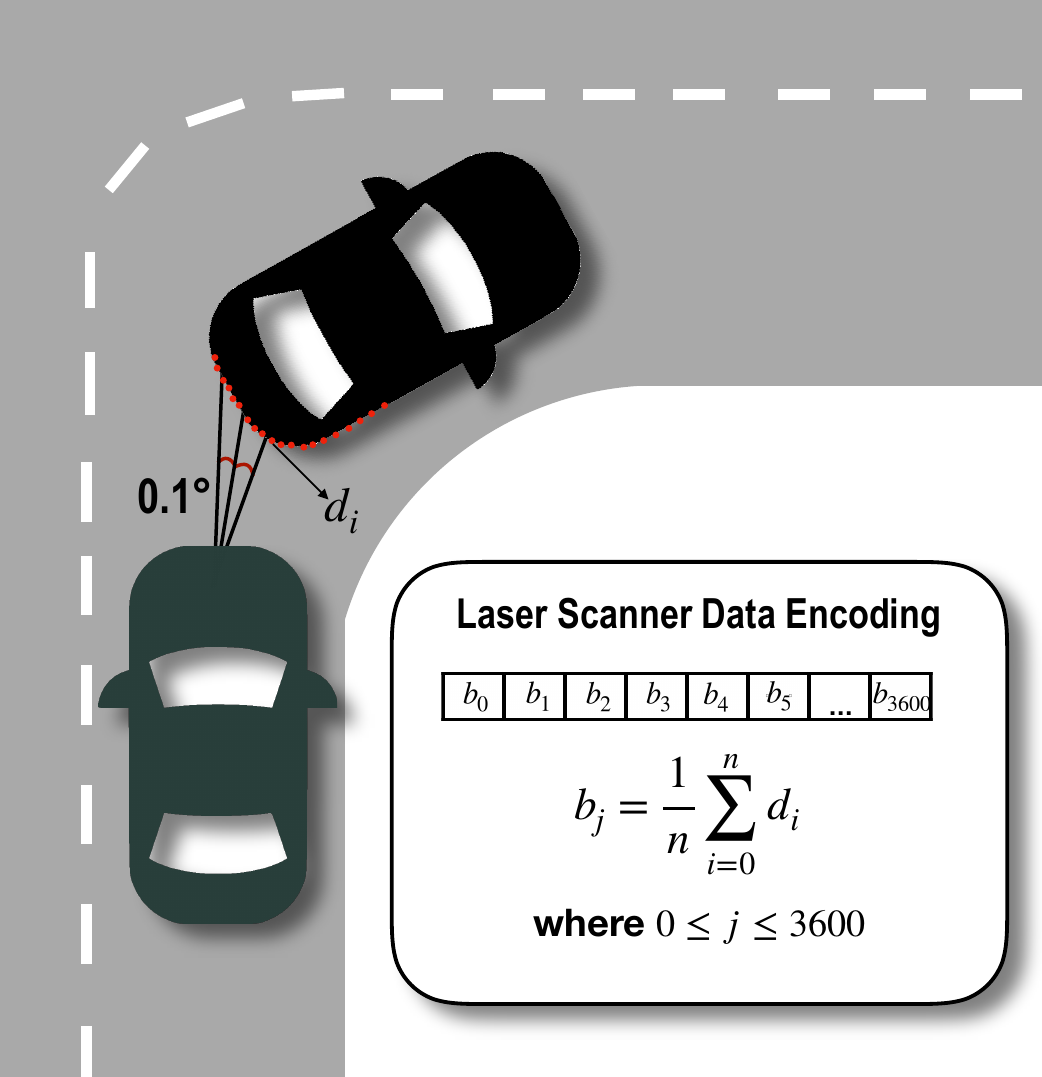}
	\caption{Data enconding for the 2D laser scanner with a $\ang{360}$ rotation range. At each time step the raw data of the laser scanner is separated into  $\ang{0.1}$ bins. The average depth of all the points in a specific bin is calculated and stored into a vector. The result is a 3601 size vector that stores the depth values for each bin.}
	\label{fig:encoding}
\end{figure}

We base our data encoding on the previous work \cite{laser2}, where the 2D laser scanner point set is encoded into a 1D vector. This can be done by first binning the raw scans into bins of resolution $\ang{0.1}$. Sequentially, since a group of points can fall into the same bin, we calculate the average depth value of this group. Finally, considering all the bins of a $\ang{360}$ rotation range, we store the depth values into a 3601 size vector, where each possible bin angle average depth is represented by the elements in the vector. This process is presented in Figure \ref{fig:encoding}. 

Once we have processed two 1D vectors from sequential scans, we concatenate them to use as input for the network. In this way, we create a form of image of size $2 x 3601$ that represents two acquisitions of the laser scanner. This format allows to use standard convolutional layers to extract the features detected by the sensor in the surrounding environment. 

\subsection{Network Architecture}

\begin{figure*}
	\centering
	\includegraphics[width=0.92\linewidth]{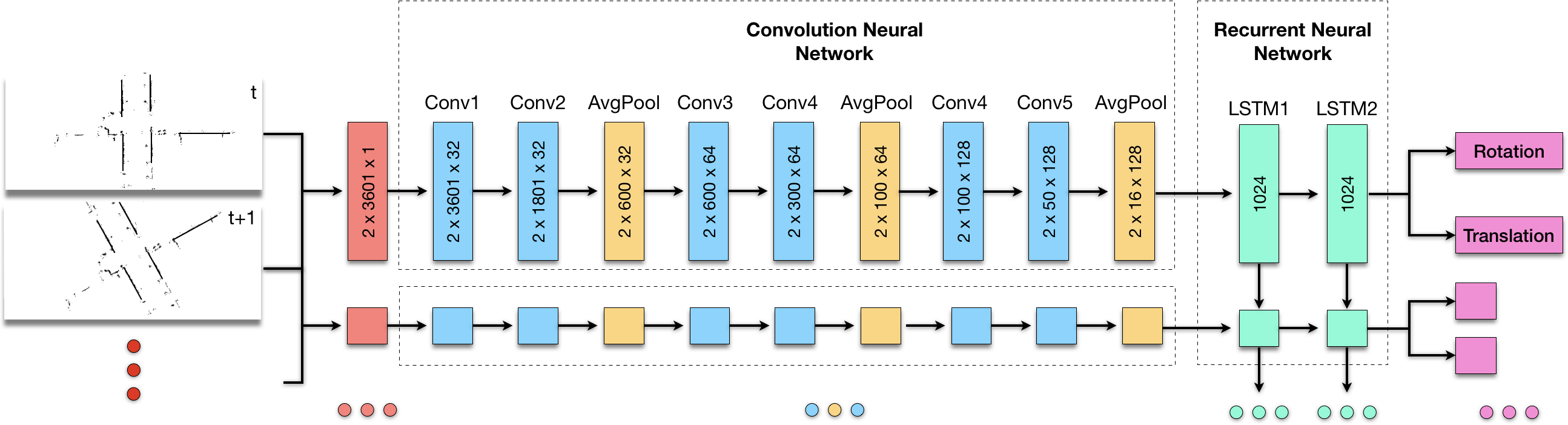}
	\caption{Architecture of the proposed RCNN. Each block of the illustration presents the size of the tensors considering that the input is two sequences of laser scanners concatenated after being encoded as presented in subsection \ref{sub.encoding}. }
	\label{fig:architecture}
\end{figure*}

The previous work on laser-based odometry estimation \cite{laser1}\cite{laser2}\cite{laser3}, for 2D and 3D sensors, only explored the use of CNNs without any deep temporal structure to estimate the local poses. We propose an architecture based on previous Visual Odometry (VO) methods, such as DeepVO \cite{deepvo}, that not only extracts the features of the input data but also estimate the possible connections among consecutive image inputs. The use of RNNs is convenient for this purpose because of its ability of modelling sequential dependencies. By adding this to our model, we aim to increase the accuracy of the pose estimation by using implicitly information that was detected by the laser scanner in previous frames. This process can be compared to graph-based approaches in classical SLAM methods \cite{graph}, where it takes a sequential of pose features and outputs a more precise estimation of these poses. 

Figure \ref{fig:architecture} presents the architecture of the proposed network. Two pre-processed laser scanner acquisitions, represented as a 1 dimension vector of size 3601, are concatenated to create the input tensor of the network. Sequentially, the tensor is fed into the sequence of 1D convolutional and average pool layers to learn the features between the two acquisitions. Then, at each new data acquisition these features are received by the RNN to estimate new poses. We represent the 2D motion by two variables, the translation (travelling distance $\Delta{d}$) and rotation $(\Delta{\theta})$ . The main goal of this process is to use the neural network to learn the features of laser scanner data while simultaneously matching them using the proposed combination of CNN and RNN. 
We inspire our network on the configuration used by DeepVO, which tries to achieve the same goal but using as input camera images. Since our data has a considerable smaller size we adapted the configuration for this purpose. This new configuration is presented in Table \ref{tab:table}. It has 6 1D convolutional layers, where each layer is followed by a rectified linear unit (ReLU) activation. Between each sequence of two convolutional layers, there is also one average pool layer. We added the pooling layers to reduce computation complexity by extracting the most important features, we tested both max and average pooling and we obtained better results applying the average pooling layer. Also, considering the size of the input and the size of features we can capture with this kind of sensor, we chose to use only kernels of size 3 after testing it with different configurations such as 5 and 7 size kernels. 

\begin{table}[H]
\center
\begin{tabular}{cccc}
\hline
\textbf{Layer} & \textbf{Kernel Size} & \textbf{Stride} & \textbf{Number of Channels} \\ \hline
Conv1                                  & 3                    & 1               & 32                          \\
Conv2                                  & 3                    & 2               & 32                          \\
Conv3                                  & 3                    & 1               & 64                          \\
\multicolumn{1}{l}{Conv4}              & 3                    & 2               & 64                          \\
\multicolumn{1}{l}{Conv5}              & 3                    & 1               & 128                          \\
\multicolumn{1}{l}{Conv6}              & 3                    & 2               & 128                          \\ \hline
\end{tabular}
\caption{Configuration of the convolutional layers in the proposed network.}
\label{tab:table}
\end{table}

After we learned the features, the output of $Conv6$ is passed to the RNN for sequential modelling. We use as our RNN, Long Short-Term Memory (LSTM) units, which are able to learn long-term dependencies \cite{lstm}. The use of stacked LSTM layers is common to learn high level representation and model complex dynamics. For this reason, we use the configuration of two LSTM layers with the hidden states of the first LSTM being used as input for the second one. The two layers are defined with 1024 hidden states. Finally, the last LSTM layer outputs two values, the rotation and translation at each time step. 

\subsection{Training}
\label{sub:training}
The goal of using RNNs is to discover temporal correlations between the sequence of laser scans. While in principle the RNN is a simple and powerful model, in practice, it can be hard to train it properly to converge to precise results \cite{diff_rnn}. For this reason, the training of the network was performed in two steps. First, we trained the sequence of CNNs separately from the RNN. The objective in the first training is to pre-train the convolutional layers using no temporal information, only considering the information obtained from the two sequential laser scans input. Once we obtained the CNN part of the network pre-trained weights, we trained the complete network as presented in Fig. \ref{fig:architecture}. 

\subsubsection{CNN Pre-Training}

To perform the pre-training, the output of the sequence of CNNs is fed to two different fully connected layers, one to estimate the rotation and another the translation. Sequentially, we train this convolutional network to learn the optimal function $g(.)$ presented in \autoref{learning_function}.

In \cite{laser3} the authors suggested that designing the network to regress the relative translation and rotation worked well only for translation, however the predicted rotation was still inaccurate. They were able to obtain better results reformulating the problem as a classification task for the rotation, and continuing as a regression one for the translation. This is possible because the range of possible rotations between consequent frames is quite reasonable. Considering this, we tested two configurations to pre-train our network, as a regression-only task, and as a regression and classification task. 

For the regression-only task, we performed the training based on the Euclidean loss between the ground truth and the estimated translation and rotation values, defining the complete loss function as follow:]
\begin{equation}
\begin{split}
    &\mathcal{L} = \mathcal{L}_{e}(\Delta{\hat{d}}, \Delta{d}) + \beta 
\: \mathcal{L}_{e}({\Delta{\hat{\theta}},\Delta{\theta}})\\
&\text{where } \mathcal{L}_e(\hat{x}, x) = \norm{{\hat{x}} - {x}}_2 
\end{split}
\label{loss_function}
\end{equation}

For the classification task, instead we used the Cross-entropy loss function to classify the angle, defining the new complete loss function for the training as:
\begin{equation}
\begin{split}
    &\mathcal{L} = \mathcal{L}_{e}(\Delta{\hat{d}}, \Delta{d}) + \beta 
\: \mathcal{L}_{c}({\Delta{\hat{\theta}},\Delta{\theta}})\\
 &\text{where } \mathcal{L}_c(x,class) = - \log \Big( \frac{\exp({x[class]})} {\sum_j{\exp(x[j])}} \Big)
\end{split}
\label{loss_classification}
\end{equation}

In \eqref{loss_function} and \eqref{loss_classification}, $\Delta{{d}}$ and $\Delta{\theta}$ are relative ground-truth translation and rotation values, and $\Delta{\hat{d}}$ and $\Delta{\hat{\theta}}$ their output of the network counterparts. We use the parameter $\beta > 0$ to balance the scale difference between the rotation and translation loss values.

Considering all the possible variation of angles between two frames, we created classes for the interval $\pm\ang{5.6}$ with $\ang{0.1}$ resolution, resulting in 112 possible classes. As indicated in \cite{laser3}, the results as a classification task for rotation were better compared to the only regression, and for this reason we used the CNN network trained as a classification for angle estimation as input for the RNN. 

\subsubsection{RCNN Training}
After initializing the weights of the CNN layers at the pre-training stage, we train the complete RCNN network. We define as the input of the RNN the output of the CNNs layers concatenated with the estimated result for rotation and translation, obtained from the pre-trained network, in a way that this could be used as a first estimation for the RCNN to refine the results.

\bgroup
\def\arraystretch{1.5}
\begin{table*}[]
\begin{tabular}{c|cc
>{\columncolor[HTML]{EFEFEF}}c c|cc}
\hline
Sequence & CNN-Regression & CNN-Classification & RCNN-Regression & RCNN-Classification & DeepVO \cite{deepvo} & 3D LiDAR CNN \cite{laser3} \\ \hline
05                         & 0.2888            & 0.2764             & \cellcolor[HTML]{EFEFEF}0.0293          & 0.2488              & 0.0262 & 0.0235              \\
07                         & 0.1281            & 0.0756             & \cellcolor[HTML]{EFEFEF}0.0218          & 0.1251              & 0.0391 & 0.0177              \\ \hline
Mean                       & 0.2084            & 0.1760             & \cellcolor[HTML]{EFEFEF}0.0255          & 0.1869              & 0.0326 & 0.0206              \\ \hline
Time (s/frame) & 0.005             & 0.005              & 0.015                                   & 0.015               & 1.0    & 0.7                 
\end{tabular}
\caption{RMSE translation drift results for the two testing sequences along with the computation time per frame without GPU acceleration. We show the difference between the use of only CNNs and RCNN, the results for both of the RCNNs are using the pre-trained CNN-Classification network. In addition, we present the error for the different training configurations presented in Subsection \ref{sub:training}. We also compare the proposed approach to two other Deep Learning odometry estimation methods, one using as the sensor a monocamera \cite{deepvo} and the second using a 3D LiDAR \cite{laser3}. However, the result presented for the method \cite{laser3} is for a training dataset, since they chose different sequences for testing.}
\label{tab:error}
\end{table*}

For the RCNN we also tested the two different configurations (regression and regression with classification). Differently, we obtained more precise results treating the entire task (rotation and translation) as a regression problem, therefore using the loss function in \eqref{loss_function}. We presume that this occurs because the estimation of the rotation as a regression was easier once we had a first estimation of the class using the pre-trained CNNs, and with the possible information obtained by the RNN from previous frames. 


\section{Results}\label{results.sec}

For validation we use the KITTI dataset \cite{kitti}, which provides several sequences in different conditions for outdoor environments. To obtain a 2D laser scanner dataset, we extracted one $\ang{360}$ layer from the Velodyne data. As mentioned before, the layer is extracted to simulate a low-cost 2D laser scanner, which could be an essential factor for the autonomous industry to maintain a reasonable price for future intelligent vehicles.

We use 10 sequences from the KITTI odometry dataset for the proposed method. In this dataset there are 11 sequences, however we eliminate the sequence 01 that consists of a trajectory mainly on a highway. During this sequence the 2D laser scanner detects almost no obstacle, making it impossible for the network to predict the odometry. Among the 10 sequences, we separate 8 for training and 2 for testing. We use for training the sequences 00, 02, 03, 04, 06, 08 and 09 and for testing the sequences 05 and 07. We chose these two sequences because they are not very long, leaving more data for the training, but they can still be challenging and present the potential of the proposed method. Additionally, during these two sequences we noticed that most of the time the simulated 2D laser scanner is able to detect obstacles, making it possible for the network to work as expected.

In order to validate our method and compare to other solutions, we calculated the drift according to the KITTI VO \cite{kitti} evaluation metrics, i.e., averaged Root Mean Square Errrors (RMSEs) of the translational error for all subsequences (100, 200,..., 800 meters). We adapt the proposed method to calculate the error only in 2D, since we only obtain 2D poses. The error score is then calculated by the mean of all subsequence errors. 

Table \ref{tab:error} presents the error score for the testing sequences 05 and 07. The difference of scores between classification and regression shows why we chose to pre-train the convolutional layers as a classification task for the angle estimation, but to treat it as a regression task when we trained the entire RCNN. These results suggest that estimating the angle as a regression task was too hard for the network to learn; however, once we had a first estimation about the class in the training of the RCNN, we were able to refine the value and obtain a more precise angle.  

\begin{figure}[H]
    \centering
    \begin{subfigure}[b]{1.0\linewidth}
    \centering
    {\includegraphics[width=0.9\linewidth]{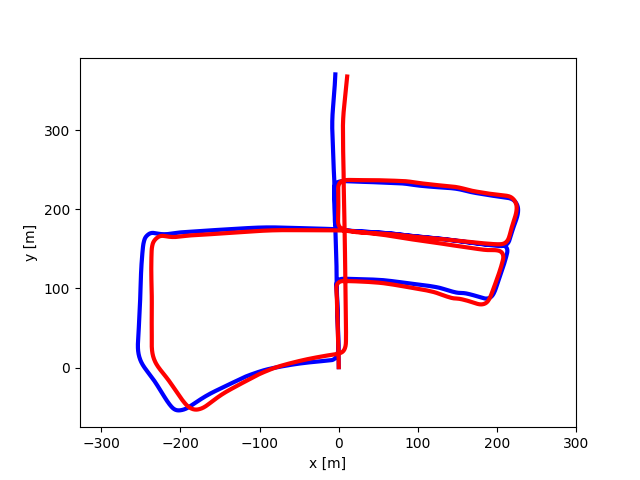} }
    \caption{Sequence 05}
    \end{subfigure}
    \qquad
    \begin{subfigure}[b]{1.0\linewidth}
    \centering
    {\includegraphics[width=0.9\linewidth]{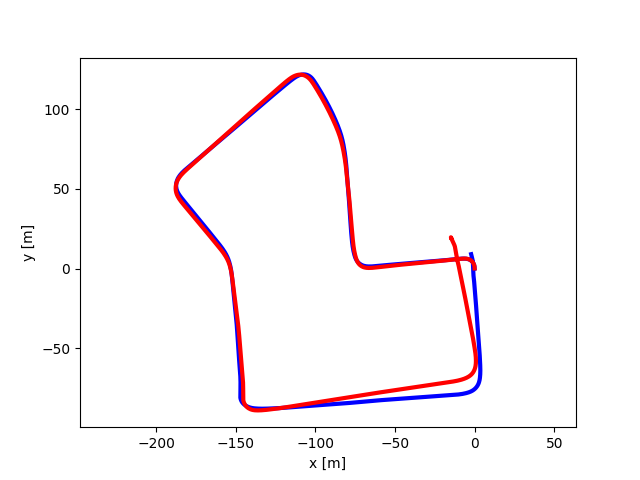} }
    \caption{Sequence 07}
    \end{subfigure}
    \caption{Trajectories of the two test sequences (05 and 07) applying the proposed method. The blue lines represent the ground truth trajectory, while in red the predicted one.}%
    \label{fig:trajectory}%
\end{figure}

\begin{figure*}
\centering
\begin{subfigure}{1.0\textwidth}
  \centering
  \includegraphics[width=1.0\linewidth]{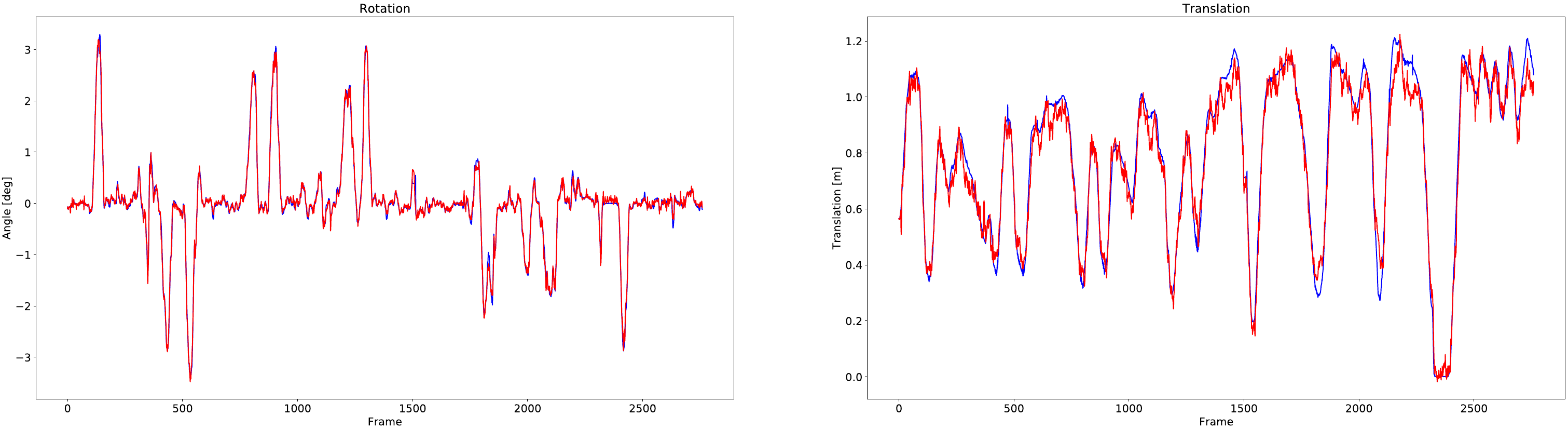}
  \caption{Sequence 05.}
  \label{fig:sub1}
\end{subfigure}%
\vspace{0.5cm}
\begin{subfigure}{1.0\textwidth}
  \centering
  \includegraphics[width=1.0\linewidth]{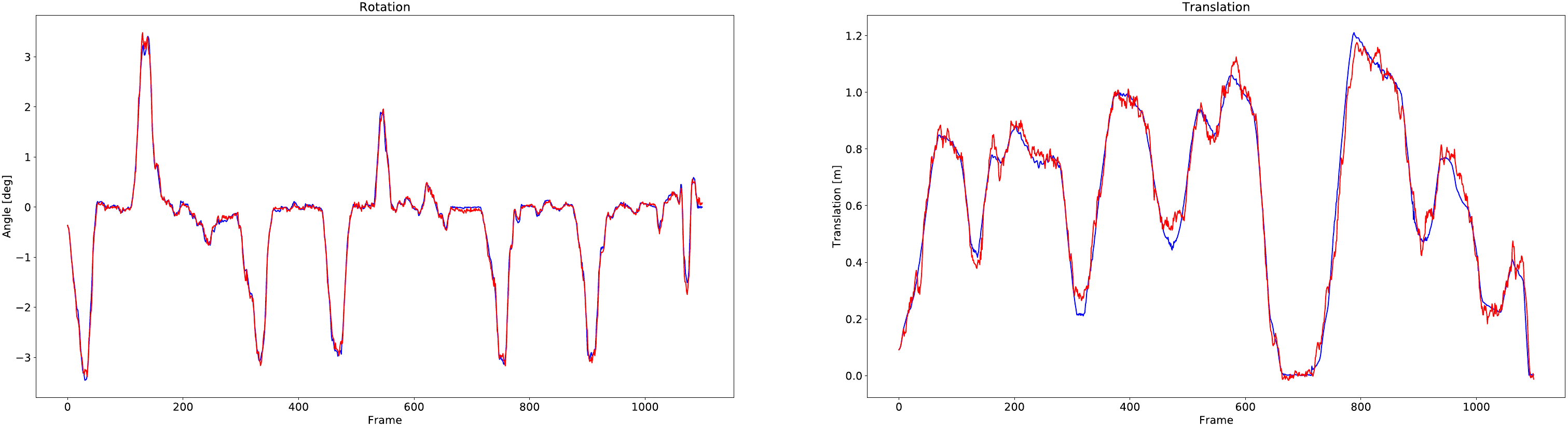}
  \caption{Sequence 07}
  \label{fig:sub2}
\end{subfigure}
\vspace{0.5cm}
\caption{Results of rotation and translation estimation for the two testing sequences. The ground truth values are presented in blue and in red the output of the network. }
\label{fig:rot_trans}
\end{figure*}

We also compare in \autoref{tab:error} the proposed network to other two Deep Learning approaches for odometry estimation: one using the mono-camera and another a 3D laser scanner. We chose these two approaches because they use the same dataset, allowing us to compare the drift results, and also their computational time is presented by the KITTI benchmark.  Our approach has better results in comparison to the network using only mono-camera images, however slightly worse results as the 3D laser scanner method. The better result in \cite{laser3} is expected, since in our experiments we are extracting only one layer of the laser scanner, which provides a significant less amount of information. In addition, as the sensor is on top of the vehicle and we always extract the most parallel layer in relation to the vehicle, there are some frames where nothing or not much is detected, causing possible wrong estimations of translation and rotation. However, it is important to mention that the 3D laser scanner approach \cite{laser3} takes 0.23s to run with GPU acceleration, while our approach takes only 0.015s per frame without using GPU acceleration (2,6 GHz Intel Core i5, Intel Iris 1536 MB), and it can be as fast as 0.001s per frame with GPU acceleration (4,0 GHz Intel Core i7, GeForce GTX 1060); i.e. a 230-fold increase in speed (GPU configuration), while obtaining only a $0.49\%$ difference in drift score and using a much cheaper sensor. The faster processing is expected since we have a considerable smaller data input, and it allows to obtain odometry estimation in real-time with simple computational resources. 

We can observe in Figure \ref{fig:trajectory} that even with the eventual errors that can occur, we can still obtain a trajectory close to the ground truth. However, since the proposed approach does not perform any sort of loop closure, one eventual large error can be accumulated over time, generating a large drift like the one we have by the end of sequence 07.  

For this reason, a better way to understand the accuracy of the proposed approach is presented in \autoref{fig:rot_trans}. It shows the odometry estimation (rotation and translation) together with the ground truth for each frame of the testing sequences. Considering these two sequences, the average rotation absolute error is 0.05 degrees, while the average translation absolute error is 0.02 meters. However, we can encounter errors up to 0.4 degrees to rotation and 0.2 meters to translation in frames where it is harder to estimate the odometry. These values present how the network can most of the time estimate accurate odometry, however there are still some difficult cases that can result in inaccurate values. 

The results show how promising is the proposed method and it could be used as a complement to traditional localization methods for intelligent vehicle or any mobile robot, when for example there are no wheel encoders or GPS signal. We can also expect that if the sensor was located in an ideal position, for example for an autonomous car as a set of 2D laser scanners around the vehicle in the level of the bumper, we could obtain even better results. It is also important to mention that we trained the network with a relatively small dataset compared to other deep learning applications, therefore the result could be improved using more sequences for training.

\section{Conclusion and Future Work}\label{conclusion.sec}

In this paper we presented a novel approach based on RCNNs to estimate the odometry using only the data of a 2D laser scanner. The combination of CNNs and RNNs allows us to achieve in real-time the extraction of scan features and learn their sequential model to obtain the localization of an intelligent vehicle. The proposed network presents that the use of 2D laser scanners can not only provide good accuracy with a low cost sensor, but also requires less computational resources to achieve real-time performance.

The results were evaluated using the KITTI odometry dataset making it possible to compare it with other Deep Learning approaches. Although the results were competitive to this type of approaches, we still do not expect that the deep learning methods could replace classic approaches at this moment, since they can still provide a better accuracy and a better understanding of the quality of the results. However, the proposed approach could be an interesting complement for classic localization estimation methods, since it can be run in real-time and could give relatively accurate values in systems where no wheel encoder data is provided or GPS signal is absent. Moreover, the proposed method shows a promising use of Neural Networks to understand the environment detected by a 2D laser scanner, since we can assume that the network learned what were the best features to match between different scans. 

In the future work, we expect that better results could be obtained after training the network with a dataset that has real 2D laser scanners located at a better position at the vehicle. Additionally, the use of more than one sensor could be explored to increase the accuracy of the results. For example, the use of a mono-camera alone was not able to get precise results, but possibly by creating a network that could perform the fusion with a 2D laser scanner these results could be improved.

\addtolength{\textheight}{-12cm}  




\end{document}